\documentclass[10pt,twocolumn,letterpaper]{article}

\usepackage{iccv}
\usepackage{times}
\usepackage{epsfig}
\usepackage{graphicx}
\usepackage{amsmath}
\usepackage{amssymb}
\usepackage{multirow}
\usepackage{tabularx}
\usepackage{algorithm}
\usepackage{algorithmic}

\usepackage{array}
\newcommand{\PreserveBackslash}[1]{\let\temp=\\#1\let\\=\temp}
\newcolumntype{C}[1]{>{\PreserveBackslash\centering}p{#1}}

\usepackage[pagebackref=true,breaklinks=true,letterpaper=true,colorlinks,bookmarks=false]{hyperref}

\makeatletter
\newcommand{\thickhline}{%
    \noalign {\ifnum 0=`}\fi \hrule height 1pt
    \futurelet \reserved@a \@xhline
}
\makeatother

\iccvfinalcopy 

\def\iccvPaperID{2798} 

\ificcvfinal\fi

\DeclareMathOperator*{\argmax}{arg\,max}

\DeclareMathOperator*{\diag}{diag}
\begin{document}

\title{FAMNet: Joint Learning of Feature, Affinity and Multi-dimensional Assignment for Online Multiple Object Tracking}

\author{Peng Chu and Haibin Ling\\
Temple University\\
Philadelphia, PA USA\\
{\tt\small \{pchu, hbling\}@temple.edu}\\
}

\maketitle

\begin{abstract}

Data association-based multiple object tracking (MOT) involves multiple separated modules processed or optimized differently, which results in complex method design and requires non-trivial tuning of parameters. In this paper, we present an end-to-end model, named \emph{FAMNet}, where Feature extraction, Affinity estimation and Multi-dimensional assignment are refined in a single network. All layers in FAMNet are designed differentiable thus can be optimized jointly to learn the discriminative features and higher-order affinity model for robust MOT, which is supervised by the loss directly from the assignment ground truth. We also integrate single object tracking technique and a dedicated target management scheme into the FAMNet-based tracking system to further recover false negatives and inhibit noisy target candidates generated by the external detector. The proposed method is evaluated on a diverse set of benchmarks including MOT2015, MOT2017, KITTI-Car and UA-DETRAC, and achieves promising performance on all of them in comparison with state-of-the-arts.
\end{abstract}


\begin{figure*}[!ht]
	\centering
	\includegraphics[width=1\linewidth]{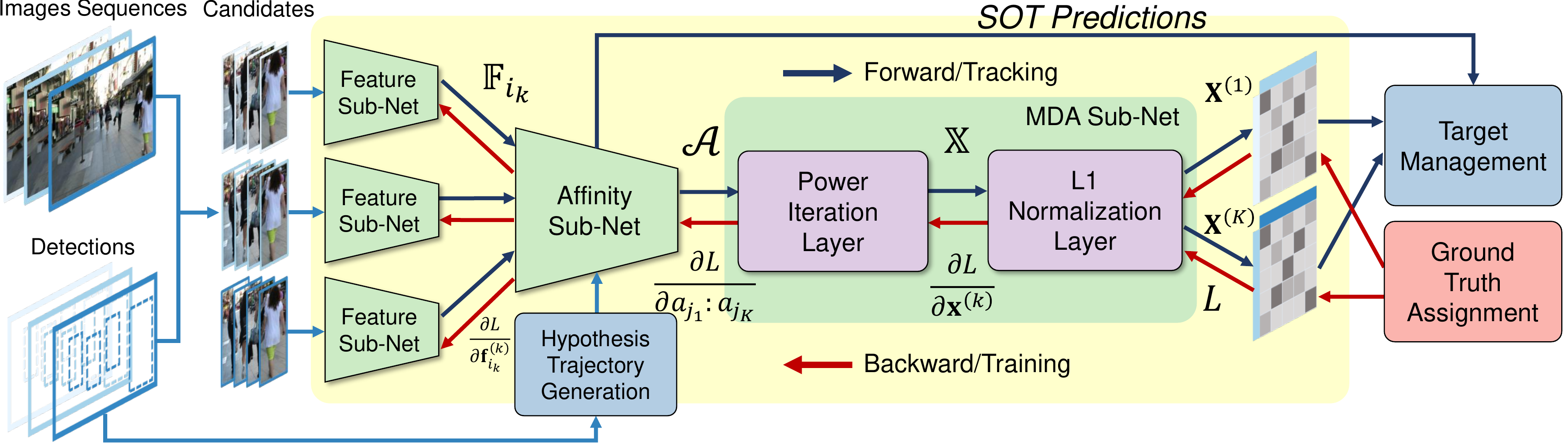}
	\caption{Overview of our FAMNet based tracking system. The sub-networks inside the yellow background consist the FAMNet. $\mathbb{F}_{i_k} $ is a set of features extracted from each frame as detailed in Sec.~\ref{subsection:affinity} and $L$ is for the total loss.}
	\label{fig:overview}
\end{figure*}

\section{Introduction}

Tracking multiple objects in video is critical for many applications, ranging from vision-based surveillance to autonomous driving. A current popular framework to solve multiple object tracking (MOT) uses the tracking-by-detection strategy where target candidates generated from an external detector are associated and connected to form the target trajectories across frames~\cite{bae2018confidence,fagot2016improving,keuper2018motion,milan2016multi,pirsiavash2011globally,wang2016tracking,xiang2015learning}. At the core of tracking-by-detection strategy lies the data association problem which is usually treated as three separate parts: \textit{feature extraction} for candidate representation, \textit{affinity metric} to evaluate the cost of each association hypothesis and \textit{association algorithm} to find the optimal association. These parts involve multiple individual data-processing steps and are optimized differently from each other, which results in a complex method design and extensive tuning parameters to adapt different target categories and tracking scenarios.

Recently, deep neural network (DNN) has been investigated intensively to learn the association cost function in a unified architecture combining both feature extraction and affinity metric~\cite{chu2017online,lan2018interacting,son2017multi}. Through training, the task and scenario prior can be automatically adapted by the candidate representation and estimation metric without manually tuning the hyper-parameters. However, the association algorithm still stands outside the network, which requires dedicated affinity samples to be manually fabricated from ground truth association for the training process. It is not guaranteed that training and inference phases share the same data distribution; consequently it may lead to the degraded generalizability of the trained model. Moreover, crowded targets, similar appearance and fast motion impose great ambiguity for the association only considering pairs of neighboring frames. Successful association requires global optimization across multiple frames, where higher-order discriminative clues such as appearance changes over time and motion context could be included. Learning the robust representation and affinity criteria without the cooperation from the association procedure in this circumstance is even more complicated.

Our objective in this paper is to formulate an end-to-end model for MOT: the Feature representation, Affinity model and Multi-dimensional assignment (MDA) are refined in a single deep network named FAMNet, which is optimized jointly to learn the task prior. In particular, feature sub-network is used to extract features for candidates on each frame, after which an affinity sub-network estimates the higher-order affinity for all association hypothesis. With the affinity, the MDA sub-network is to optimize globally and obtain the optimal assignments. By all layers in FAMNet designed differentiable, the feature and affinity sub-network can be trained directly referring to the assignment ground truth. To realize it, we make the following novelties to the FAMNet and its based tracking system:
\begin{itemize}
\item\vspace{-2mm} We design an affinity sub-network that fuses discriminative higher-order appearance and motion information into the affinity estimation.
\item\vspace{-2mm} We propose an MDA sub-network, in which a modified rank-1 tensor approximation power iteration is designed differentiable and adapted for the deep learning architecture.
\item\vspace{-2mm} We integrate single object tracking into the data association-based MOT. Detections and tracking predictions are merged and selected optimally through MDA to construct the target trajectories.
\item\vspace{-2mm} We employ a target management scheme where a dedicated CNN network is used to refine the target bounding box to eliminate the noised candidates generated by external detector.
\end{itemize}

\vspace{-2mm}
To show the effectiveness of the proposed approach, it is evaluated on the popular multiple pedestrian and vehicle tracking challenge benchmarks including MOT2015, MOT2017, KITTI-Car and UA-DETRAC. Our results show promising performance in comparison with other published works.

\section{Related Work}
\label{sec:relate_work}

Multiple object tracking (MOT) has been an active research area for decades, and many methods have been investigated for this topic. Recently, the most popular framework for MOT is the tracking-by-detection. Traditional methods primarily focus on solving the data association problem using such as Hungarian algorithm~\cite{bewley2016simple,fang2018recurrent,huang2008robust}, network flow~\cite{dehghan2015target,zamir2012gmcp,zhang2008global} and multiple hypotheses tracking~\cite{chen2017enhancing,kim2015multiple} on various of affinity estimation schemes. Higher-order affinity provides the global and discriminative information that is not available in pairwise association. In order to utilize it, MOT is usually treated as the MDA problem. Collins~\cite{collins2012multitarget} proposes a block ICM-like method for the MDA to incorporate higher-order motion model. The method iteratively solves bipartite assignments alternatively while keeping other assignments fixed. In~\cite{shi2019rank}, MDA is formulated as the rank-1 tensor approximation problem where a dedicated power iteration with unite $\ell1$ normalization is proposed to find the optimal solution. Our work is closely related to the MDA formulation, especially \cite{shi2019rank}.

Recently, deep learning is explored with increasing popularity in MOT with great success. Most recent solutions rely on it as a powerful discriminative technique~\cite{bae2018confidence,leal2016learning,son2017multi,zhu2018online}. Tang et al.~\cite{tang2017multiple} propose to use DNN based Re-ID techniques for affinity estimations. They include lift edges that connect two candidates spanning multiple frames to capture the long-term affinity. In~\cite{sadeghian2017tracking}, recurrent neural networks (RNN) and long short-term memory (LSTM) is adapted to model the higher-order discriminative clue. Those methods learn the networks in a separate process with the manually fabricated affinity training samples.

Some recent works have gone further to tentatively solve MOT in an entirely end-to-end fashion. Ondruska and Posner~\cite{ondruska2016deep} introduce the RNN for the task to estimate the candidate state. Although this work is demonstrated on the synthetic sensor data and no explicit data association is applied, it firstly shows the efficacy of using RNN  for an end-to-end solution. Milan et al.~\cite{milan2017online} propose an RNN-LSTM based online framework to integrate both motion affinity estimation and bipartite association into the deep learning network. They use LSTM to solve the data association target by target at each frame where the constrains in data association are not explicitly built into the network but learned from training data. For both works, only the occupancy status of targets are considered, the informative appearance clue is not utilized. Different from their methods, we propose an MDA sub-network which handles both the data association and the constrains, and our affinity fuses both the appearance and motion clue for better discriminability.

\section{Overview}
\label{sec:overview}
In this section, we first formulate the multiple object tracking (MOT) problem as a multi-dimensional assignment (MDA) form, and then provide an overview of our FAMNet-based tracking system (overview in Fig.~\ref{fig:overview}).

\subsection{Problem Formulation}
Following the notation in \cite{shi2019rank}, the input for MOT is denoted by $\mathbb{O} = \{\mathbb{O}^{(k)}\}_{k=0}^K$, which contains $K + 1$ target candidate sets from $K + 1$ frames. For frame $k$, $\mathbb{O}^{(k)} = \{\mathbf{o}^{(k)}_{i_k}\}_{i_k=1}^{I_k}$ is the set of $I_k$ candidates to be matched or associated, where $\mathbf{o}^{(k)}_{i_k}$ represents the status of the candidate such as its center coordinate on the image frame.

With the input candidate set $\mathbb{O}$, MOT is to find a multi-dimensional association that maximizes the overall affinity subject to the association constrains. In detail, $c_{i_0: i_K} \doteq c_{i_0i_1...i_K} \ge 0$ denotes affinity for one possible association, or in term of MOT, one hypothesis trajectory $\mathbf{t}_{i_0:i_K} $ composed by candidates $\big\{\mathbf{o}^{(0)}_{i_0}, \mathbf{o}^{(1)}_{i_1}, ..., \mathbf{o}^{(K)}_{i_K}\big\}$. We use $z_{i_0: i_K} \doteq z_{i_0i_1...i_K}$ to indicate whether a hypothesis trajectory is true ($z_{i_0: i_K} = 1$) or not ($z_{i_0: i_K} = 0$). If we further denote tensor $\mathcal{C} =(c_{i_0: i_K}) $ and $ \mathcal{Z} = (z_{i_0: i_K})$, the MOT can be formulated as following MDA problem to solve $ \mathcal{Z}$ based on $\mathcal{C}$:
\begin{equation}
\label{eq:obj}
\argmax_{\{z_{i_0: i_K}\}} \sum_{i_0: i_K}c_{i_0: i_K}z_{i_0: i_K}  = \argmax_{\mathcal{Z}}\| \mathcal{C} \circ \mathcal{Z} \|_1,
\end{equation}
\begin{equation}
\label{eq:st}
\mathrm{s.t.}
\begin{cases}
\sum\limits_{i_0: i_K/\{i_k\}}z_{i_0: i_K} = 1, &\forall k = 0, 1, ..., K \\
z_{i_0: i_K} \in \{0, 1\} , &\forall i_k = 1, 2, ..., I_k \\
\end{cases}
\end{equation}
where $\circ$ denotes the element-wise product, $\| \cdot \|_1$ is the matrix 1-norm, and $\sum_{i_0: i_K/\{i_k\}}$ stands for summation over all subscripts $i_0: i_K$ except for $i_k$.

To solve Eq.~\ref{eq:obj}, we follow the Rank-1 Tensor Approximation (R1TA) framework~\cite{shi2019rank}. The multi-dimensional assignments $\mathcal{Z}$ are first decomposed as the product of a serials of local assignments $\mathbf{X}^{(k)} = (x_{i_{k-1}i_k}^{(k)})$ which represent the assignments between candidates in adjacent frames, \ie $\mathbb{O}^{(k-1)}$ and $\mathbb{O}^{(k)}$. If we further rewrite the local assignment matrix into vector form $\mathbf{x}^{(k)} = (x_{j_k}^{(k)})$ where $j_k = (i_{k-1} - 1) \times I_k + i_{k}$,\footnote{For  notational convenience, the same symbol is used for  elements in $\mathbf{X}^{(k)}$ and $ \mathbf{x}^{(k)}$ with double subscripts and a single subscript respectively.} optimization problem defined in Eq.~\ref{eq:obj} can be rewritten as follow:
\begin{equation}
\label{eq:obj_re}
\argmax_{\mathbb{X}} \mathcal{A} \times_1 \mathbf{x}^{(1)}\times_2 \mathbf{x}^{(2)} \cdots \times_K \mathbf{x}^{(K)},
\end{equation}
where $\mathcal{A}=(a_{j_1:j_K})$ is the reshaped affinity tensor from the $(K+1)$-th order $\mathcal{C}$ tensor to a $K$-th order tensor following the rules defined in~\cite{shi2019rank} and $\times_k$ is the $k$-$mode$ tensor product, $\mathbb{X} = \big\{\mathbf{x}^{(1)}, \mathbf{x}^{(2)}, ..., \mathbf{x}^{(K)} \big\}$ is the set of local assignment vectors which we are optimizing for.

\subsection{Architecture Overview and Tracking Pipeline}

For each association batch, the FAMNet based tracking system takes the $K+1$ image frames and corresponding detections provided by an external detector as input. Detection candidates are first used to generate the hypothesis trajectories. Image patches of target candidates together with the trajectory hypothesis are passed into FAMNet to compute the set of local assignments as shown in Fig.~\ref{fig:overview}. Inside FAMNet, features of candidate patches are extracted through a feature sub-network. The affinity sub-network then calculates the affinity for all hypothesis trajectories on those features to form the affinity tensor as described in Sec.~\ref{subsection:affinity}. With the affinity tensor, the optimal multi-dimension assignments are estimated by the MDA sub-network as explained in Sec.~\ref{subsection:power_iter} and \ref{subsection:l1_norm}.

During training, the assignment ground truth is directly compared with the network output to compute the loss. The loss signal then back-propagates throughout the network to the feature and affinity sub-networks for learning, which is illustrated as the red paths in Fig.~\ref{fig:overview} and detailed in Sec.~\ref{subsection:train}. In tracking phase, the output assignments together with single object tracking (SOT) predictions are used to update the trajectories of tracked targets through the target management scheme as described in Sec.~\ref{subsection:tracking} and \ref{subsection:target_mgr}.

We design our method in the online tracking framework intended for more casual applications. Under the constant velocity assumption, three frames are the minimum temporal span to calculate the motion affinity. Therefore, in the rest of paper, $K = 2$ with two frames overlapping between association batches is used as to balance the computation cost and the sufficient depth of association to include the higher-order discriminative clues.

\section{FAMNet}
\label{sec:method}
\begin{figure*}[!ht]
	\centering
	\includegraphics[width=1\linewidth]{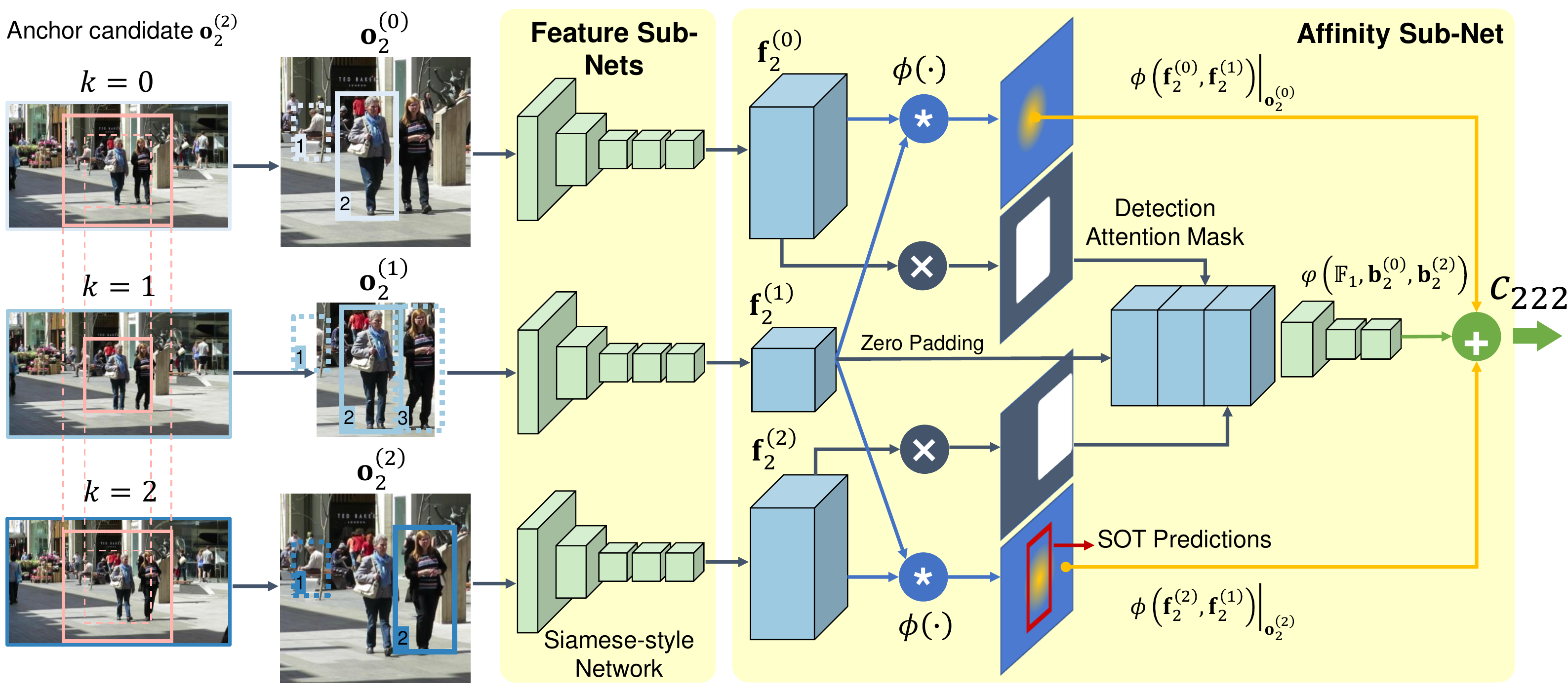}
	\caption{Illustration of the feature and affinity sub-networks for $K=2$. It shows an example to calculate the affinity $c_{222}$ for the hypothesis trajectory $\mathbf{t}_{222} $. The red bounding boxes in the first column of images indicate the locations on the image frame for the patch in the second column. The blue bounding boxes in the second column illustrate the detection candidates. Best viewed in color.}
	\label{fig:affinity}
\end{figure*}
\subsection{Affinity Sub-Network}
\label{subsection:affinity}

The affinity sub-network takes the features of candidates and hypothesis trajectories as input, and generates the affinity tensor as output.

For each association batch, the feature sub-network, which is a Siamese-style network, is first used to extract spatially aligned feature for the candidates from all frames in the batch. The candidates in the middle frame of the batch are treated as \textit{anchor candidates}, and the middle frame is referred as \textit{anchor frame}. E.g., for $K = 2$, the anchor frame refers to the frame $k = 1$. For each anchor candidate $\mathbf{o}_{i_1}^{(1)}$, $K + 1$ features are extracted respectively from the $K + 1$ frames, denoted as $\mathbb{F}_{i_1} = \{\mathbf{f}_{i_1}^{(0)}, \mathbf{f}_{i_1}^{(1)}, \mathbf{f}_{i_1}^{(2)}\}$. These features are all centered at the same location of $\mathbf{o}_{i_1}^{(1)}$ on frame $k = 1$, as illustrated on the left of Fig.~\ref{fig:affinity}. This way, features in $\mathbb{F}_{i_1}$ share the same coordinate origin thus can encode motion clue when concatenated along a channel. Spatial dimension of $\mathbf{f}_{i_1}^{(1)}$ is determined by the bounding box of $\mathbf{o}_{i_1}^{(1)}$, while others are the multiples of $\mathbf{f}_{i_1}^{(1)}$ in order to include enough candidates on adjacent frames into the same view.   Note that, hypothesis trajectories sharing the same anchor candidate have the same set of $\mathbb{F}_{i_k}$. Therefore, $(K+1)I_k$ features are extracted for each association batch.

Two levels of affinities are calculated for each hypothesis trajectory using the extracted feature set, as shown in Fig.~\ref{fig:affinity}. In detail, the affinity tensor is calculated as following:
\begin{equation}
\label{eq:affinity}
c_{i_0i_1i_2}=
\begin{cases}
\begin{aligned}[l]
&\varphi\big(\mathbb{F}_{i_1}, \mathbf{b}_{i_0}^{(0)}, \mathbf{b}_{i_2}^{(2)} \big) \\
+&\phi\big(\mathbf{f}_{i_1}^{(0)}\bigl\vert_{\mathbf{o}_{i_0}^{(0)}}, \mathbf{f}_{i_1}^{(1)}\big) \\
+&\phi\big(\mathbf{f}_{i_1}^{(2)}\bigl\vert_{\mathbf{o}_{i_2}^{(2)}}, \mathbf{f}_{i_1}^{(1)}\big),
\end{aligned}
 & \mathbf{t}_{i_0i_1i_2} \in \mathbb{T}\\
0 , &\mathbf{t}_{i_0i_1i_2} \notin \mathbb{T}, \\
\end{cases}
\end{equation}
where $\mathbb{T}$ is the set of valid hypothesis trajectories, $\mathbf{b}_{i_k}^{(k)}$ is the bounding box associated with $\mathbf{o}_{i_k}^{(k)}$, $\phi(\cdot)$ calculates the pair-wise affinity and $\varphi(\cdot)$ evaluates the long-term affinity of a hypothesis trajectory, $ \mathbf{f}_{i_1}^{(0)}\vert_{\mathbf{o}_{i_0}^{(0)}}$ is a spatial subset of $\mathbf{f}_{i_1}^{(0)}$ centered at $\mathbf{o}_{i_0}^{(0)}$ with the same spatial dimension as $\mathbf{f}_{i_1}^{(1)}$. The actual center coordinates of $\mathbf{o}_{i_0}^{(0)}$ in $\mathbf{f}_{i_1}^{(0)}$ need to be converted accordingly. We use $\mathbf{o}_{i_0}^{(0)}$ here for convenience.

For pair-wise affinity, the cross-correlation operation is used, such as
\begin{equation}
\label{eq:tracking}
\phi\big(\mathbf{f}_{i_1}^{(0)}\bigl\vert_{\mathbf{o}_{i_0}^{(0)}}, \mathbf{f}_{i_1}^{(1)}\big)
  =\mathbf{f}_{i_1}^{(1)}\ast\mathbf{f}_{i_1}^{(0)}\bigl\vert_{\mathbf{o}_{i_0}^{(0)}}
  = \phi\big(\mathbf{f}_{i_1}^{(0)}, \mathbf{f}_{i_1}^{(1)}\big)\bigl\vert_{\mathbf{o}_{i_0}^{(0)}},
\end{equation}
where $*$ is the convolution operation.  Due to the fact that hypothesis trajectories sharing the same anchor candidates have the same set of $\mathbb{F}_{i_1}$, we can calculate $\mathbf{f}_{i_1}^{(1)} * \mathbf{f}_{i_1}^{(0)}$ first then take the value at $\mathbf{o}_{i_0}^{(0)}$ from the cross-correlation result, which is referred as $\phi(\cdot)\vert_{\mathbf{o}_{i_0}^{(0)}}$.

We use convolutional neural network (CNN) with spatial attention to evaluate the higher-order affinity of hypothesis trajectory. 
For this purpose, features $\mathbf{f}_{i_1}^{(0)}$ and $\mathbf{f}_{i_1}^{(2)}$ are multiplied with spatial masks generated from $\mathbf{b}_{i_0}^{(0)}$ and $\mathbf{b}_{i_2}^{(2)}$ as shown in Fig.~\ref{fig:affinity}. In particular, we create a binary mask of the same spatial size with $\mathbf{f}_{i_1}^{(0)}$ or $\mathbf{f}_{i_1}^{(2)}$ for each candidate. Inside each mask, the region within $\mathbf{b}_{i_k}^{(k)}$ is set to 1 otherwise is 0. Each time, the actual position of $\mathbf{b}_{i_k}^{(k)}$ in the mask is converted from the image frame to the coordinates centered at anchor candidates. After encoding the spatial-temporal information, features in $\mathbb{F}_{i_1}$ are concatenated along channel to form the input of a CNN to estimate the long-term affinity. The final affinity for a hypothesis trajectory is the summation of two levels of affinity according to Eq.~\ref{eq:affinity}.

\subsection{R1TA Power Iteration Layer}
\label{subsection:power_iter}
With the affinity tensor, we use R1TA power iteration to estimate the set of optimal assignments in Eq.~\ref{eq:obj_re}. Solving the global optimum for MDA usually requires NP-hard probing. A sub-optimal approximation is usually guaranteed by a power iteration algorithm which can be expressed in the pure mathematic format.

In order to fit this process into the deep network framework, we adapt a different iteration scheme than the one in~\cite{shi2019rank} where the row/column $\ell1$ normalization is applied to $\mathbb{X}$ after each iteration to enforce the constrain defined in Eq.~\ref{eq:st}. The tensor power iteration and row/column normalization are separated into two independent layers in our design. It avoids cumulating too deep operations in a single layer and alleviate the potential gradient vanishing. Downside of this scheme is that we could not expect the same convergence property as in \cite{shi2019rank}. However, benefited from the end-to-end training, it can be compensated by the learned more discriminative feature and affinity metric.

In detail, the optimal solution to Eq.~\ref{eq:obj_re} subject to Eq.~\ref{eq:st} is approximated iteratively by:\\
\textbf{Forward pass}. At the $(n+1)$-th iteration, the elements in $\mathbf{x}^{(1)(n + 1)}$ is calculated from\footnote{The second superscript indicates the round of iteration. Moreover, derivation in this subsection is on $x_{j_1}^{(1)}$, but is the same for other $x_{j_k}^{(k)}$.}
\begin{equation}
x_{j_1}^{(1)(n + 1)} = \frac{x_{j_1}^{(1)(n)}}{C^{(n)}}\hspace{-3mm}\sum_{j_1:j_K/\{j_1\}}\hspace{-3mm}a_{j_1:j_K}x_{j_2}^{(2)(n)}\cdots x_{j_K}^{(K)(n)},
\end{equation}
where $C^{(n)}= \sum_{j_1:j_K}a_{j_1:j_K}x_{j_1}^{(1)(n)}x_{j_2}^{(2)(n)}\cdots x_{j_K}^{(K)(n)}$ is the $\ell1$ normalization factor. At initialization, elements in all local assignment vectors $\mathbf{x}^{(k)(0)}$ are set to 1.\\
\textbf{Backward pass}. The R1TA power iteration layer computes the loss gradient of affinity tensor $\mathcal{A}$, denoted by $\partial L/\partial a_{j_1:j_K}$, as backward output. The input of the backward pass is the loss gradient of all local assignment vectors at the last iteration, e.g. $\partial L / \partial \mathbf{x}^{(k)(N)}$, where $N$ is the total number of iterations performed in the forward pass. The gradient output is calculated as follow:
\begin{equation}
\label{eq:pi_grad}
\begin{aligned}
\frac{\partial L}{\partial a_{j_1:j_K}} =
&\sum_n \frac{x_{j_1}^{(1)(n)}x_{j_2}^{(2)(n)}\cdots x_{j_K}^{(K)(n)}}{C^{(n)}} \\
&\times \sum_k \big(\mathbf{i}_{j_k}^{(k)} - \mathbf{x}^{(k)(n + 1)}\big)^{\top}\frac{\partial L}{\partial \mathbf{x}^{(k)(n+1)}},
\end{aligned}
\end{equation}
where $\mathbf{i}_{j_k}^{(k)}$ is a unit vector of the same dimension as $\mathbf{x}^{(k)}$ and has elements equal to 1 only at $j_k$ and otherwise 0. In order to calculate Eq.~\ref{eq:pi_grad} for all iterations, the loss gradients of assignment vectors at each iteration are also needed, which follows:
\begin{equation}
\begin{aligned}
\hspace{-2.5mm}\frac{\partial L}{\partial x_{j_1}^{(1)(n)}} = \frac{x_{j_1}^{(1)(n +1)}}{x_{j_1}^{(1)(n)}} \Big[ \big(\mathbf{i}_{j_1}^{(1)} - \mathbf{x}^{(1)(n + 1)}\big)^{\top}\hspace{-1.5mm}\frac{\partial L}{\partial \mathbf{x}^{(1)(n+1)}} \\
-\sum_{k \neq 1}\big(\mathbf{x}^{(1)(n + 1)}\big)^{\top}\hspace{-1.5mm}\frac{\partial L}{\partial \mathbf{x}^{(1)(n+1)}} \Big] + \frac{\sum_{k \neq 1} h_{j_1}^{(k)(n)}}{C^{(n)}},
\end{aligned}
\end{equation}
where $h_{j_1}^{(k)(n)}$ can be calculated as, \eg for $k=2$, $h_{j_1}^{(2)(n)} = \sum_{l_1:l_K/\{l_1\}} a_{j_1l_2\cdots l_k}x_{l_3}^{(3)(n)}\cdots x_{l_K}^{(K)(n)}\frac{\partial L}{\partial x_{l_2}^{(2)(n)}}$.

\begin{figure}
	\centering
	\includegraphics[width=0.9\linewidth]{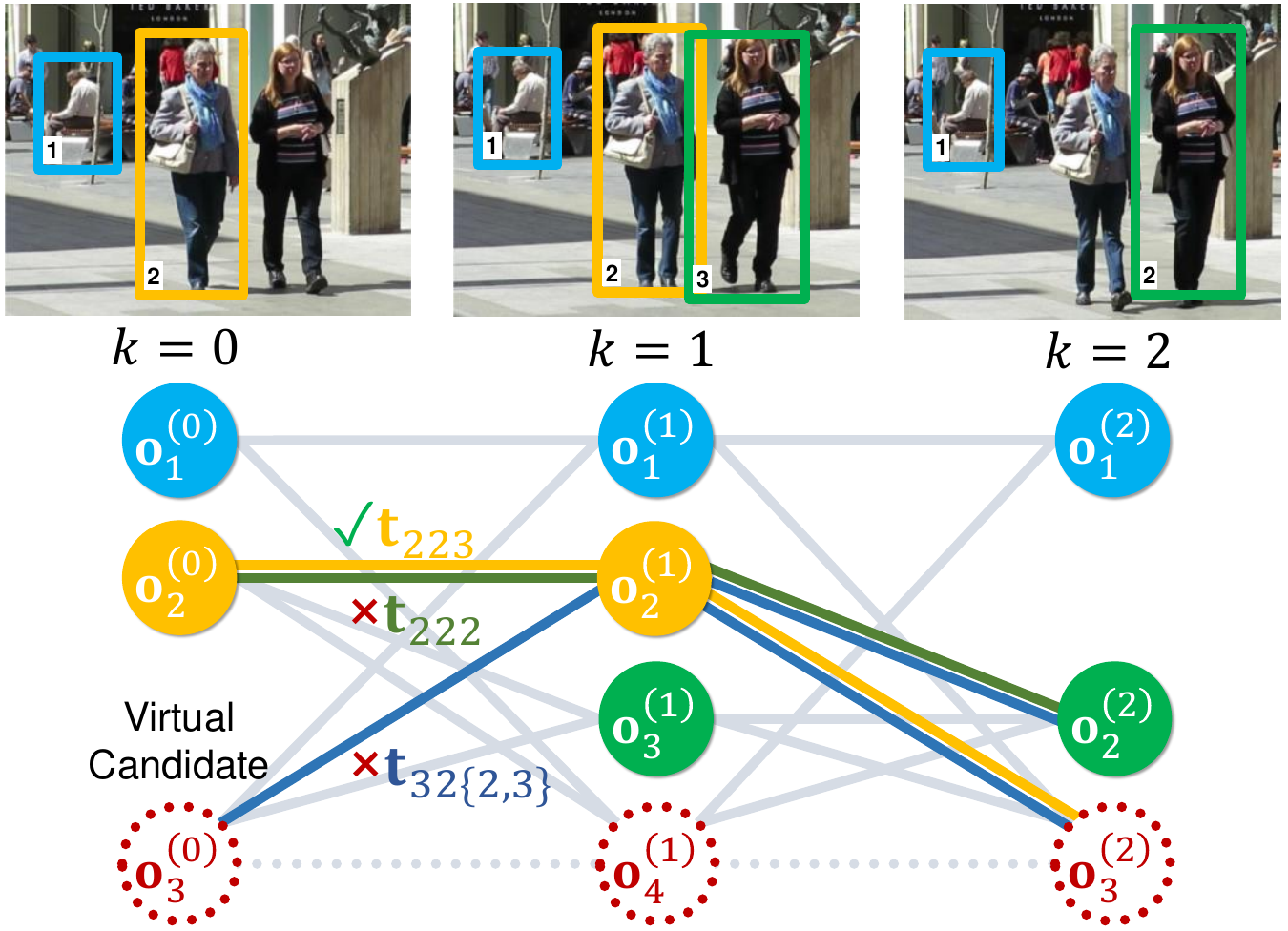}
	\caption{Hypothesis trajectory generation. It shows all hypothesis trajectories passing through candidate $\mathbf{o}_2^{(1)}$. The color of nodes indicates the ground truth associations.}
	\label{fig:hypothesis}
\end{figure}

\subsection{$\ell1$ Normalization Layer}
\label{subsection:l1_norm}
To satisfy the constrains defined in Eq.~\ref{eq:st} required by MDA, row/column $\ell1$ normalization is applied to the result $\mathbb{X}$. The assignment vectors from the R1TA power iteration layer are reshaped back to their matrix form such that
$\mathbf{X}^{(k)} = (x_{i_{k-1}i_k}^{(k)}) \in \mathbb{R}^{I_{k-1}\times I_{k}}$. Then, the $\ell1$ normalization is performed row and column alternatively through multiple iterations.\\
\textbf{Forward pass}. For each pair of iterations, we start from row normalization. In the $(n + 1)$ and $(n + 2)$-th iteration:
\begin{equation}
\begin{aligned}
\mathbf{X}^{(k)(n + 1)} &= \big[ \mathbf{X}^{(k)(n)}\mathbf{1}_{I_k} \big]^{-1}\mathbf{X}^{(k)(n)} \\
\mathbf{X}^{(k)(n + 2)} &= \mathbf{X}^{(k)(n + 1)}\big[\mathbf{1}_{I_{k-1}}^{\top} \mathbf{X}^{(k)(n + 1)}\big]^{-1},
\end{aligned}
\end{equation}
where $\mathbf{1}_{I_k}$ is a vector of $\mathbb{R}^{I_k}$ with all elements being 1, $[\mathbf{x}]$ here and below represents the diagonal matrix with $\mathbf{x}$ as diagonal elements.\\
\textbf{Backward pass}. Given a starting gradient $\partial L / \partial \mathbf{X}^{(k)(n + 2)}$, we iteratively compute the gradients as: 
\begin{equation}
\begin{aligned}
    \frac{\partial L}{\partial \mathbf{X}^{(k)(n + 1)}} &= \frac{\partial L}{\partial \mathbf{X}^{(k)(n + 2)}}\big[\mathbf{1}_{I_{k-1}}^{\top} \mathbf{X}^{(k)(n + 2)}\big]^{-1} -\mathbf{1}_{I_{k-1}}\\
    \cdot \diag\Big(\big[&\mathbf{1}_{I_{k-1}}^{\top} \mathbf{X}^{(k)(n + 2)}\big]^{-2} \big( \mathbf{X}^{(k)(n + 2)}\big)^{\top}  \hspace{-1.5mm}\frac{\partial L}{\partial \mathbf{X}^{(k)(n + 2)}}\Big)^{\top} \\
    \frac{\partial L}{\partial \mathbf{X}^{(k)(n)}} &= \big[\mathbf{X}^{(k)(n + 1)} \mathbf{1}_{I_{k}} \big]^{-1} \hspace{-1.5mm}\frac{\partial L}{\partial \mathbf{X}^{(k)(n + 1)}}\\
    - \diag\hspace{-.4mm}\Big(\hspace{-.4mm}\big[&\mathbf{X}^{(k)(n + 1)}\mathbf{1}_{I_{k}}\big]^{\hspace{-.4mm}-2}  \hspace{-1.5mm} \frac{\partial L}{\partial \mathbf{X}^{(k)(n + 1)}} \big(\mathbf{X}^{(k)(n + 1)}\big)^{\top} \hspace{-.4mm}\Big)\mathbf{1}_{I_{k}}^{\top}.
\end{aligned} \nonumber
\end{equation}

Our $\ell1$ normalization layer is similar to but differs from that in \cite{zanfir2018deep}, in that our implementation allows partial row/column normalization to handle real and virtual candidates differently as detailed in Sec.~\ref{subsection:tracking}.

\subsection{Training}
\label{subsection:train}

During the training, the total loss $L$ is measured by the binary cross entropy between all predicted assignments $\big(x_{i_{k-1}i_{k}}^{(k)} \in [0, 1 ] \big)$ and assignment ground truth  $\big(\bar{x}_{i_{k-1}i_{k}}^{(k)}\in\{0, 1\} \big)$ , which is written as
\begin{equation}
L \hspace{-.7mm}= \hspace{-.987mm}\sum_k \hspace{-.987mm}\sum_{i_{k-1}i_k} \hspace{-.97mm}\bar{x}_{i_{k-1}i_{k}}^{(k)} \hspace{-.7mm} \log x_{i_{k-1}i_{k}}^{(k)}
 \hspace{-.7mm}+ (1 - \bar{x}_{i_{k-1}i_{k}}^{(k)}\hspace{-.7mm})\log(1 - x_{i_{k-1}i_{k}}^{(k)}\hspace{-.7mm}). \nonumber
\end{equation}
With the total loss, gradients are calculated throughout the network back to the affinity and feature sub-networks.

We use each association batch containing $K+1$ frames as one mini-batch during training and tracking. For each batch, $K+1$ consecutive frames and target candidates on each frame provided by an external detector serve as the input. The candidate set is first used to generate the hypothesis trajectories $\{\mathbf{t}_{i_0:i_K}\}$. We set a bound for the hypothesis trajectory generation where two candidates from two consecutive frames can be connected only when they are spatially close to each other and have the similar bounding box size. We set adaptive thresholds for this strategy. In detail, if a candidate cannot connect with any candidate at a threshold, it re-searches using degraded thresholds for the possible connection. This strategy allows the connections for both fast and slow movement targets, and meanwhile rejects naive false positive connections. Hypothesis trajectories are generated greedily by iterating through all valid connections to form trajectories starting with candidates in $\mathbb{O}^{(0)}$ and terminating in $\mathbb{O}^{(K)}$. Generated trajectories are sorted by their anchor candidates and together with image frames fed into our networks for training and tracking.

\begin{algorithm} [!t]
	\small
	\caption{\small Target Management}
	\begin{algorithmic}[1]
		\STATE Input: Assignment matrix $\mathbf{X}^{(k)} = (x_{i_{k-1}i_k}^{(k)}) \in \mathbb{R}^{I_{k-1}\times I_{k}}$. \\
		\STATE Output: Tracked target trajectories.\\
		\STATE Discrete $\mathbf{X}^{(k)}$ using graph multicut~\cite{keuper2015efficient}.\\
		\FOR {$i_{k - 1}=1,\dots,I_{k-1} $}
		\IF {$\mathbf{o}_{i_{k-1}}^{(k-1)}$ not tracked \\
			\textbf{and} $\mathrm{CNN_{BBE}}\big(\mathbf{b}_{i_{k-1}}^{(k - 1)}\big) > 0.5$}
		\STATE add new target trajectory starting with $\mathbf{b}_{i_{k-1}}^{(k - 1)}$
		\ENDIF
		
		\IF {$\mathbf{o}_{i_{k-1}}^{(k-1)}$ assigned to a real candidate, \textit{e.g.} $\mathbf{o}_{j_k}^{(k)}$}
		\IF {IoU$\big(\mathbf{p}_{i_{k-1}}^{(k-1)}$, $\mathbf{b}_{j_k}^{(k)}\big) < t_{\text{dif}}$\\ \textbf{and} $\mathrm{CNN_{BBE}}\big(\mathbf{b}_j^{(k)}\big) < 0.5$}
		\STATE update target trajectory of $\mathbf{o}_{i_{k-1}}^{(k-1)}$ using $\mathbf{p}_{i_{k-1}}^{(k - 1)}$
		\STATE continue
		\ENDIF
		\STATE update target trajectory of $\mathbf{o}_{i_{k-1}}^{(k-1)}$ using $\mathbf{b}_{j_k}^{(k)}$
		\ELSE
		\STATE \(\triangleright\) $\mathbf{b}_F$ is the bounding box of image frame.
		\IF {IoU$\big(\mathbf{p}_{i_{k-1}}^{(k-1)}, \mathbf{b}_F\big) < t_{\text{exit}}$  }
		
		\STATE exit target trajectory of $\mathbf{o}_{i_{k-1}}^{(k-1)}$
		\ELSE
		\STATE update target trajectory of $\mathbf{o}_{i_{k-1}}^{(k-1)}$ using $\mathbf{p}_{i_{k-1}}^{(k - 1)}$
		\ENDIF
		\ENDIF
		\ENDFOR
	\end{algorithmic}
	\label{alg:target_mgr}
\end{algorithm}

\subsection{Tracking by Integrating Detection and SOT}
\label{subsection:tracking}
In the tracking phase, predictions using SOT techniques are included to recover missing candidates from the external detector. We add a virtual candidate to each candidate set to represent missing candidates and allow it to connect with any candidate in consecutive frames as shown in Fig.~\ref{fig:hypothesis}. Both real and virtual candidates are used to generate trajectory hypothesis. When calculating affinity, we choose the location maximizing the affinity in Eq.~\ref{eq:tracking} as the center of the virtual candidates for each anchor candidate such that
\vspace{-1mm}
\begin{equation}
\label{eq:sot}
\vspace{-2.5mm}\argmax_{\hat{\mathbf{o}}_{I_2 + 1}^{(2)}} \phi\big(\mathbf{f}_{i_1}^{(2)}, \mathbf{f}_{i_1}^{(1)}\big)
=\big( \mathbf{f}_{i_1}^{(1)}\ast\mathbf{f}_{i_1}^{(2)}\big) \bigl\vert_{\hat{\mathbf{o}}_{I_2 + 1}^{(2)}}.
\end{equation}
Therefore, if an anchor candidate misses its detection in consecutive frame, it will connect with the virtual candidate which represents the location most similar to it in that consecutive frame, or, in terms of SOT, its tracking prediction. Each anchor candidate may have a different location predicted by SOT. We use $\hat{\mathbf{o}}_{I_2 + 1}^{(2)}$ in Eq.~\ref{eq:sot} to refer to the virtual candidate, its center coordinates may vary on different anchor candidates.

To prevent MDA from always choosing the virtual candidates since their affinity are no smaller than any real candidate, a coefficient $\alpha \in (0, 1)$ is used to scale down their affinity. The virtual candidates are handled specially in the $\ell1$ normalization layer:
for the row (column) in $\mathbf{X}^{(k)(n + 1)}$ representing a virtual candidate, only column (row) $\ell1$ normalization is applied.
This way, each real candidate can be assigned to only one candidate in consecutive frame including the virtual ones, while a virtual candidate can be assigned to multiple candidates. During optimization, if the affinities of real candidates are smaller than that of the virtual candidates, which represent tracking predictions, the anchor candidates will be automatically associated with the tracking predication. This way, our tracking system integrates the detection and SOT naturally.

\subsection{Target Management}
\label{subsection:target_mgr}
On receiving the assignment results, target management handles target entering, exiting and updating. In assignment results, if multiple anchor candidates choose to associate with a virtual candidate, new candidates will be added into candidate sets accordingly. For a virtual candidate not associated with any anchor candidate in assignment results, it will be dropped from the candidate set. Furthermore, if the virtual candidate associated with an anchor candidate in this batch appears as an anchor candidate in the next batch, the appearance feature of the anchor candidate is reused in the next batch in case that the missing detection is caused by occlusion. This SOT process will continue until a confident real candidate is associated.

For anchor candidates associated with real candidates, we train a CNN network to further refine their bounding boxes. During MDA, associations are made mainly based on the object center of each candidate. Most MOT tasks concern the actual bounding box enclosure of targets. When an anchor candidate is assigned to a real candidate, two bounding boxes will be associated with it: one from the real candidate itself denoted by $\mathbf{b}_{i_k}^{(k)}$ and the other from the SOT predication of the anchor candidate $\mathbf{p}_{i_{k-1}}^{(k - 1)}$. If the bounding box Intersection over Union (IoU) between $\mathbf{b}_{i_k}^{(k)}$ and $\mathbf{p}_{i_{k-1}}^{(k - 1)}$ is smaller than a threshold $t_{\text{dif}}$, a CNN is used to evaluate the quality of $\mathbf{b}_{i_k}^{(k)}$. In detail, a CNN-based binary classifier $\mathrm{CNN_{BBE}}$ is trained to decide whether a bounding box has IoU larger than a threshold, \eg 0.5, with the category target. 
The detailed procedure of the target management is listed in Alg.~\ref{alg:target_mgr}.

\begin{table}[!t]
	\footnotesize
	\begin{center}
		\caption{Tracking Performance on the MOT2015 benchmark test set. Best in bold.}
		\label{table:res15}
		\begin{tabular}{@{\hskip 1mm}c@{\hskip 1mm}|@{\hskip 1.5mm}c@{\hskip 1.5mm}c@{\hskip 1.5mm}c@{\hskip 1.5mm}c@{\hskip 1.5mm}c@{\hskip 1.5mm}c@{\hskip 1.5mm}c@{\hskip 1.5mm}c@{\hskip 1.5mm}}
			\hline\thickhline
			& Method & MOTA & MOTP & MT & ML$\downarrow$ & FP$\downarrow$ & FN$\downarrow$ & IDS$\downarrow$\\
			\hline
			
			\multirow{5}{*}{\rotatebox{90}{\textbf{Offline}}}
			&CEM \cite{milan2014continuous}& 19.3& 70.7& 8.5\%& 46.5\%& 14180 &34591 &813\\
			
			&R1TA \cite{shi2019rank}& 24.3 & 68.2& 5.5\%& 46.6\%& 6664& 38582 &1271\\
			
			& SCNN \cite{leal2016learning}& 29.0& 71.2& 8.5\%& 48.4\%& \textbf{5160} &37798& 639\\
			
			& DAM \cite{kim2015multiple}& 32.4&71.8&16.0\% &43.8\% &	9064 &	32060&	\textbf{435}\\
			
			& JMC \cite{keuper2018motion}&\textbf{35.6}&\textbf{71.9}&\textbf{23.2\%} &\textbf{39.3\%} &10580 &\textbf{28508} &457\\
			\cline{1-9}
			
			\multirow{5}{*}{\rotatebox{90}{\textbf{Online}}}
			&RNN \cite{milan2017online}& 19.0	&71.0&	5.5\%&	45.6\% &11578&36706&	1490\\
			
			&oICF \cite{kieritz2016online}& 27.1&70.0	&6.4\% &48.7\% &7594&36757&	\textbf{454}\\
			
			& SCEA \cite{hong2016online}& 29.1	&71.1	&	8.9\% &	47.3\% &6060&	36912	&604\\
			
			& AP \cite{chen2017online}& 38.5& \textbf{71.3}& 8.7\%& 37.4\%& \textbf{4005} &33203 &586\\
			
			& proposed & \textbf{40.6} & 71.1 & \textbf{12.5\%} & \textbf{34.4\%} &  4678 & \textbf{31018}	& 778 \\
			
			\hline\thickhline
		\end{tabular}
		
	\end{center}
	
\end{table}

\begin{table}[!t]
	\footnotesize
	\begin{center}
		\caption{Tracking Performance on the MOT2017 benchmark test set. Best in bold.}
		\label{table:res17}
		\begin{tabular}{@{\hskip 1mm}c@{\hskip 1mm}|@{\hskip 1mm}c@{\hskip 0mm}c@{\hskip 1mm}c@{\hskip 1.2mm}c@{\hskip 1.2mm}c@{\hskip 1.2mm}c@{\hskip 1.2mm}c@{\hskip 1.2mm}c@{\hskip 1mm}}
			\hline\thickhline
			& Method & MOTA & MOTP & MT & ML$\downarrow$ & FP$\downarrow$ & FN$\downarrow$ & IDS$\downarrow$\\
			
			\hline
			\multirow{4}{*}{\rotatebox{90}{\textbf{Offline}}}
			& IOU17 \cite{bochinski2017high}&45.5&	76.9&15.7\% &	40.5\% &	19993& 281643&	5988\\
			&bLSTM \cite{kim2018multi}& 47.5&77.5&	18.2\% &	41.7\% &	25981&268042&	2069\\
			
			&TLMHT \cite{sheng2018iterative}& 50.6 &	\textbf{77.6}&	17.6 \% &	43.4\% &	\textbf{22213}&	255030	&\textbf{1407}\\
			
			& jCC \cite{tang2017multiple}&\textbf{51.2} &	75.9 & 	\textbf{20.9\%} &	\textbf{37.0\%} &	25937&	\textbf{247822} &	1802\\
			\cline{1-9}
			
			\multirow{4}{*}{\rotatebox{90}{\textbf{Online}}}
			&GMPHD \cite{kutschbach2017sequential}& 39.6	&74.5&	8.8\% &	43.3 \% &	50903 &	284228 &	5811\\						
			& DMAN \cite{zhu2018online}& 48.2 &	75.7&\textbf{19.3\%} &	38.3 \% &	26218&	263608&	\textbf{2194}\\
			& MOTDT \cite{long2018tracking} & 50.9 & \textbf{76.6} & 17.5\% & 35.7\% & 24069 & \textbf{250768} & 2474 \\
			& proposed & \textbf{52.0 } & 76.5   & 19.1\% &	\textbf{33.4\%}&	\textbf{14138}&	253616&	3072 \\
			\hline\thickhline
		\end{tabular}
		
	\end{center}
	
\end{table}

\section{Experiment}
\label{sec:exp}


We conduct experiments on four popular MOT datasets: MOT2015~\cite{MOTChallenge2015} and MOT2017~\cite{MOT16} for pedestrian tracking, KITTI-Car~\cite{geiger2012we} and UA-DETRAC~\cite{wen2015ua} for vehicle tracking. All datasets are provided with referred detections from real detectors.

\subsection{Experiment Setting}

The proposed approach is implemented in PyTorch and runs on a desktop with CPU of 6 cores@3.60GHz and a Titan X GPU. We adapt the SiamFC proposed in~\cite{sadeghian2017tracking} as our feature sub-network and use their weights as pre-trained model which is trained on the ILSVRC15 dataset for object detection in video. The CNN $\varphi(\cdot)$ to estimate long-term affinity is constructed with three convolutional layers to map the concatenated spatial aligned feature into affinity score. A ResNet-101 with binary output is adapted for $\mathrm{CNN_{BBE}}$, the pre-trained weights from MaskRCNN~\cite{he2017mask} on COCO dataset is used for initialization. Proposed method runs average 0.6 fps on MOT2017 dataset in tracking phase.

For each test sequence in MOT2015 and MOT2017, following their protocol, one or more similar sequences in the training set are used to train a different set of FAMNet and $\mathrm{CNN_{BBE}}$ to best adapt the scenario prior. Sequences in the KITTI and UA-DETRAC dataset are all recorded in a similar setting, therefore all training sequences in their datasets are used together to train one set of networks for all test sequences. To train the FAMNet, ground truth bounding boxes are used as input target candidates. Therefore, no virtual candidate or SOT process is enabled during training. When training the $\mathrm{CNN_{BBE}}$, the training samples are collected from both the external detection and ground truth bounding box after random shift and scale. The IoU of bounding boxes with ground truth larger than 0.5 are selected as positive samples while smaller than 0.4 are for negative samples.

To evaluate the performance of the proposed method, the widely accepted CLEAR MOT metrics \cite{bernardin2008evaluating} are reported, which include multiple object tracking precision (MOTP) and multiple object tracking accuracy (MOTA) that combines false positives (FP), false negatives (FN) and the identity switches (IDS). Additionally, we also report the percentage of mostly tracked targets (MT), the percentage of mostly lost targets (ML).

\subsection{Evaluation Results}

\noindent\textbf{MOT2015.}
MOT2015~\cite{MOTChallenge2015} contains 11 different indoor and outdoor scenes of public places with pedestrians as the objects of interest, where camera motion, camera angle and imaging condition vary greatly. The dataset provides detections generated by the ACF-based detector~\cite{dollar2014fast}. The numerical results on its test set are reported in Tab.~\ref{table:res15}. Our approach achieves clearly the state-of-the-art performance. In particular, our method achieves better performance in most metrics than the RNN based end-to-end online methods due to our discriminative higher-order affinity and the optimization method adapted. Our method also surpasses the same R1TA-based method which is with hand-crafted features and affinity metrics.

\vspace{1.5mm}\noindent\textbf{MOT2017.}
Similar to MOT2015, MOT2017~\cite{MOT16} contains seven different sequences in both training and test datasets but with higher average target density (31.8 vs 10.6 on the test set), thus is more challenging. MOT2017 also focuses on evaluating the tracker performance on different detection quality. It provides three different detection inputs from DPM~\cite{felzenszwalb2010object}, Faster-RCNN~\cite{ren2015faster} and SDP~\cite{yang2016exploit}, ranked in ascending order by AP. We train seven different sets of networks according to different scenes, without further fitting on the different detections. The numerical results are reported in Tab.~\ref{table:res17}. The performance of our method is better than or on par with other published state-of-the-art methods.

\vspace{1.5mm}\noindent\textbf{KITTI-Car.}
The KITTI dataset~\cite{geiger2012we} contains 21 video sequences in the training set and 29 in the test set for multiple vehicle tracking in street view, where videos are recorded through a camera mounting in front of a moving vehicle. Referred detections from the regionlet~\cite{wang2013regionlets} detector are used in our experiment. The numerical results on the dataset of our method along with other methods using the same detections are summarized in Tab.~\ref{table:kitti}. Our method again surpasses the hand-crafted feature-based R1TA method, despite the fact that it uses a much larger association batch for off-line tracking. 
It is worth mentioning that motion affinity plays a more importance role in KITTI than in MOT2015 and MOT2017, since both targets and camera move faster and more regularly in KITTI.

\vspace{1.5mm}\noindent\textbf{UA-DETRAC.}
UA-DETRAC dataset~\cite{wen2015ua} is another multiple vehicle tracking dataset with 60 sequences for training and 40 sequences for testing. All sequences are recorded with static camera at a lift-up position near different drive ways in various of weather conditions. We use referred detection from CompACT~\cite{cai2015learning} detector in our experiment. UA-DETRAC reports the average of each MOT metric from a serials of results using different detection confidence thresholds (from 0 to 1.0 with 0.1 step). Comparison with other methods using the same detections are reported in Tab.~\ref{table:ua}. Proposed method achieves state-of-the-art performance among the published works. Our method also surpasses the IOU tracker which is an offline method and using a private detector.

\begin{table}[!t]
	\footnotesize
	\begin{center}
		\caption{Tracking Performance on the KITTI-Car benchmark test set. Best in bold.}
		\label{table:kitti}
		\begin{tabular}{@{\hskip 1.2mm}c@{\hskip 1.2mm}|@{\hskip 1.2mm}c@{\hskip 1.2mm}c@{\hskip 1.2mm}c@{\hskip 1.2mm}c@{\hskip 1.2mm}c@{\hskip 1.2mm}c@{\hskip 1.2mm}c@{\hskip 1.2mm}c@{\hskip 1.2mm}}
			\hline\thickhline
			& Method & MOTA & MOTP & MT & ML$\downarrow$ & FP$\downarrow$ & FN$\downarrow$ & IDS$\downarrow$ \\
			
			\hline
			\multirow{4}{*}{\rotatebox{90}{\textbf{Offline}}}
			& DCO-X \cite{milan2013detection}& 68.1 &	78.9&	37.5\% &	14.1\% &	2588	&8063&	318 \\
			& R1TA \cite{shi2019rank}& 71.2 &	79.2&	47.9 \% &	11.7\% &	1915&	7579	&418 \\												
			& LP-SSVM \cite{wang2017learning}& 77.6 &	77.8&	56.3\% &	\textbf{8.5\%} &	1239&	\textbf{6393}&	62\\
			& NOMT \cite{choi2015near}&\textbf{78.1} &	\textbf{79.5} & 	\textbf{57.2\%} &	13.2\% &	\textbf{1061}&	6421 &	\textbf{31} \\	
			\cline{1-9}
			
			\multirow{4}{*}{\rotatebox{90}{\textbf{Online}}}
			&RMOT \cite{yoon2015bayesian}& 65.8	&75.4&	40.2\% &	9.7 \% &	4148 &	7396 &	209 \\
			
			& mbodSSP \cite{lenz2015followme}& 72.7  &	78.8	&	48.8\% &	\textbf{8.7\%} &	1918&7360&\textbf{114}	 \\
			
			& CIWT \cite{osep2017combined}& 75.4 &	\textbf{79.4}&49.9\% &	10.3 \% &	954 &	7345 &	165  \\
			
			& proposed & \textbf{77.1} & 78.8   & \textbf{51.4}\% &	8.9\%&	\textbf{760}&	\textbf{6998}&	123  \\
			
			\hline\thickhline
		\end{tabular}
		
	\end{center}
	
\end{table}

\begin{table}[!t]
	\footnotesize
	\begin{center}
		\caption{Tracking Performance on the UA-DETRAC benchmark test set. All results are averaged over different input detection confidence thresholds. Best in bold.}
		\label{table:ua}
		\begin{tabular}{@{\hskip 1.8mm}c@{\hskip 1.8mm}c@{\hskip 1.8mm}c@{\hskip 1.8mm}c@{\hskip 1.8mm}c@{\hskip 1.8mm}c@{\hskip 1.8mm}c@{\hskip 1.8mm}c@{\hskip 1.8mm}}
			\hline\thickhline
			Method & MOTA & MOTP & MT & ML$\downarrow$ & FP$\downarrow$ & FN$\downarrow$ & IDS$\downarrow$ \\
			\hline
			CEM~\cite{milan2014continuous} & 5.1 &	35.2&	3.0\% &	35.3\% & \textbf{12341} & 260390& \textbf{267} \\
			$\text{H}^2\text{T}$~\cite{wen2014multiple} & 12.4 & 35.7 & 14.8 \% & 19.4\%& 51765 & 173899 & 852 \\
			CMOT~\cite{bae2018confidence} & 12.6 & 36.1 & 16.1\% &	18.6\% & 57886&	167111&	285\\
			GOG~\cite{pirsiavash2011globally} & 14.2 &	\textbf{37.0} & 	13.9\% &	19.9\% & 32093 & 180184 & 3335 \\
			$^{\dagger}$IOU~\cite{bochinski2017high} & 19.4 & 28.9 & \textbf{17.7\%} & 18.4 \% & 14796 & 171806 & 2311\\
			proposed & \textbf{19.8} & 36.7 & 17.1\% & \textbf{18.2\%} & 14989 & \textbf{164433} & 617\\
			\hline\thickhline
		\end{tabular}
		$^{\dagger}$~Private detector used.
	\end{center}
	
\end{table}

\subsection{Ablation Study}

\begin{table}[!t]
	\footnotesize
	\begin{center}
		\caption{Ablation study on sequences from MOT2015.}
		\label{table:ablation}
		\begin{tabular}{c|cccc}
			
			\hline\thickhline
			Method & MOTA& FP$\downarrow$ & FN$\downarrow$ & IDS$\downarrow$  \\
			\hline
			No training & 35.5 & 240 & 4799 & 202\\
			Training from scratch & 44.1 & 281 & 4160 & 97  \\
			Without $ \mathrm{CNN_{BBE}}$ & 40.5 & 518 & 4227 & 87  \\
			Without SOT & 42.0 & \textbf{200} & 4412 & 99\\
			\hline
			Fine-tuning (proposed) & \textbf{45.2} & 259 & \textbf{4105} & \textbf{87}  \\
			\hline\thickhline			
			
		\end{tabular}
	\end{center}
\end{table}

We justify the effectiveness of different modules in proposed method through ablation study as shown in Tab.~\ref{table:ablation}. 
We conduct the study using the sequences ETH-Pedcross2 and ETH-Sunnyday for testing and ETH-Bahnhof for training. All sequences are from the training set of MOT2015.
We start from FAMNet with randomly initialized weights whose tracking performance is referred as ``No training" in Tab.~\ref{table:ablation}. Then the network is trained on sequence ETH-Bahnhof. ``Training from scratch" stands for the results in this scheme. Training with the limited MOT sequences may lead to overfitting of the feature and affinity sub-network. To increase the generalizability and further boost the performance, we use the weights trained on the ILSVRC15 dataset as initialization then perform fine-tuning on the MOT sequence, which is referred as ``Fine-tuning'' and is the scheme used in other experiments in this paper. ``Without $ \mathrm{CNN_{BBE}}$'' shows the configuration where detection score is used for bounding box quality estimation instead of the dedicated $\mathrm{CNN_{BBE}}$. Target management without $\mathrm{CNN_{BBE}}$ cannot efficiently prevent FPs merging into the tracking results. ``Without SOT'' in Tab.~\ref{table:ablation} stands for the case that only detections from external detector are used for association, no SOT prediction is included. By contrast, in the proposed solution, though SOT predictions introduce some FPs, it recovers much more missing candidates and reduces FN greatly. 

\section{Conclusion}
\label{sec:conclusion}

In this paper we proposed a novel deep architecture for MOT, which learns jointly, in an end-to-end fashion, features and high-order affinity directly from the ground truth trajectories. During tracking, predictions from SOT and a dedicated target management are include to further boost tracking robustness. Experiments on the MOT2015, MOT2017, KITTI-Car and UA-DETRAC datasets clearly show the effectiveness of proposed approach.

\end{document}